\documentclass[conference]{IEEEtran}

\usepackage{microtype}
\usepackage{graphicx}
\usepackage{subfigure}
\usepackage{booktabs} 

\usepackage{hyperref}

\usepackage{amsmath}
\usepackage{amssymb}
\usepackage{mathtools}
\usepackage{amsthm}
\usepackage{float}
\usepackage{algorithm}
\usepackage{algorithmic}
\usepackage{cite}
\usepackage[square, comma, sort&compress, numbers]{natbib}

\begin{document}

	\title{Symbolic Expression Transformer: \\A Computer Vision Approach for Symbolic Regression}
	
	\author{\IEEEauthorblockN{Jiachen Li}
		\IEEEauthorblockA{Department of Automation\\ Shanghai Jiao Tong University\\ Shanghai, China, 200240}
	\and \IEEEauthorblockN{Ye Yuan}
		\IEEEauthorblockA{Department of Automation\\ Shanghai Jiao Tong University\\ Shanghai, China, 200240\\ $yuanye\_auto@sjtu.edu.cn$}
	\and \IEEEauthorblockN{Hon-Bin Shen}
		\IEEEauthorblockA{Department of Automation\\ Shanghai Jiao Tong University\\ Shanghai, China, 200240\\$hbshen@sjtu.edu.cn$}}

	\maketitle

	\begin{abstract}
		Symbolic Regression (SR) is a type of regression analysis to automatically find the mathematical expression that best fits the data. Currently, SR still basically relies on various searching strategies so that a sample-specific model is required to be optimized for every expression, which significantly limits the model's generalization and efficiency. Inspired by the fact that human beings can infer a mathematical expression based on the curve of it, we propose Symbolic Expression Transformer (SET), a sample-agnostic model from the perspective of computer vision for SR. Specifically, the collected data is represented as images and an image caption model is employed for translating images to symbolic expressions. A large-scale dataset without overlap between training and testing sets in the image domain is released. Our results demonstrate the effectiveness of SET and suggest the promising direction of image-based model for solving the challenging SR problem.
		
	\end{abstract}
	
	\section{Introduction}
	\label{Introduction}
	Discovering the mathematical expressions between variables from collected data is a common concern in the history of various scientific areas. Symbolic Regression (SR) searches for a suitable structure and corresponding parameters to construct an explicit mathematical model that can best fit the observed data \cite{augusto2000symbolic}. Given a dataset ${({In}_i,\ {Out}_i)}_i$, where ${In}_i\in R^n$ and ${Out}_i\in R$, SR looks for the function $f(\cdot):\ R^n\ \rightarrow\ R$ to minimize the loss over all data points i.e., $\min_f {\sum_{i}{({f({In}_i)-{Out}_i)}^2}}$.     
	
	\begin{figure*}[t]
		\vskip 0.2in
		\begin{center}
			\centerline{\includegraphics[width=2\columnwidth]{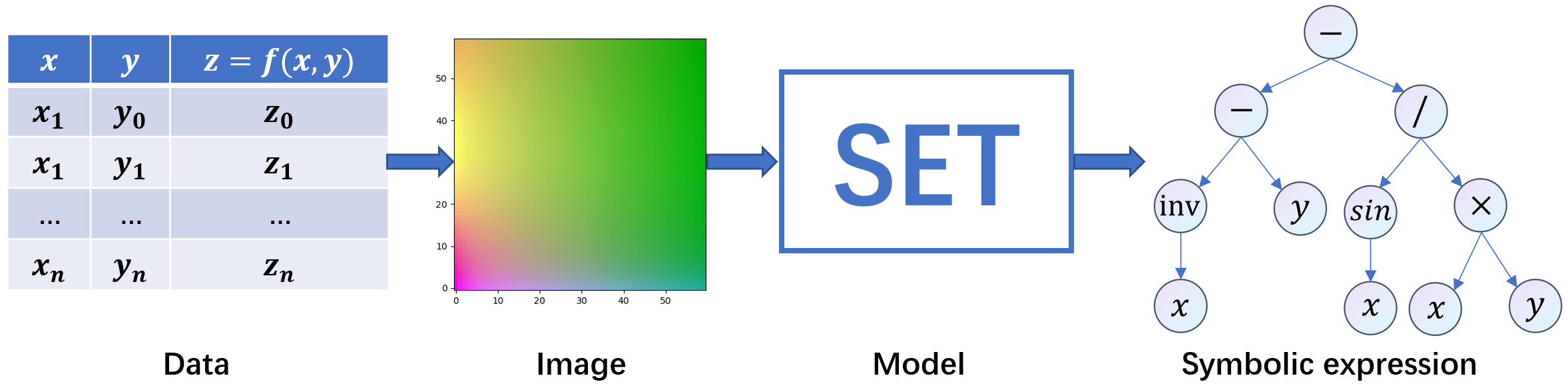}}
			\caption{The overview of SET. The collected data and the corresponding mathematical expressions are represented as input images and output symbolic sequences respectively, so that SR is modeled as an image caption task in SET.}
			\label{Overview}
		\end{center}
		\vskip -0.2in
	\end{figure*}
	
	Previous SR approaches are mainly based on searching strategies. Evolutionary algorithms, especially genetic programming (GP) methods, are widely utilized in traditional approaches \cite{back2018evolutionary, dubvcakova2011eureqa, haeri2017statistical, koza1992programming, schmidt2009distilling, uy2011semantically}. Recent advances in machine learning prompted neural networks to be applied to SR. AI Feynman \cite{udrescu2020ai} decomposes a complex SR problem to a serious of simpler ones before applying brute-force symbolic search, by employing neural networks to identify the simplifying properties such like multiplicative separability and translational symmetry. GrammarVAE \cite{kusner2017grammar} was proposed to train a variational autoencoder for obtaining the latent space representation of discrete data. Sahoo et al. \cite{sahoo2018learning} proposed a special kind of neural networks where the activation functions are symbolic operators for addressing SR. Deep symbolic regression (DSR) \cite{petersen2021deep} applied deep reinforcement learning to generate symbolic expressions and proposed a risk-seeking strategy for better exploring the searching space. Additionally, Bayes symbolic regression (BSR) \cite{jin2019bayesian} was proposed to fit SR under a Bayesian framework, and Neat-GP \cite{trujillo2016neat} optimized the standard GP approaches based on Neuro Evolution of Augmenting Topologies (NEAT). From the perspective of computer vision (CV), DeSTrOI \cite{xing2021automated} was proposed for symbolic operator identification, which aims at predicting the significance of each mathematical operator to reduce the searching space for downstream SR tasks. \cite{biggio2021neural} and \cite{kamienny2022end} employed transformer-based models in SR by training them with synthetic datasets.

	Several issues limit further applications of current approaches: (1) The absence of large-scale benchmark datasets for comprehensive evaluation. Most existing methods are only benchmark on no more than hundreds of expressions (Table \ref{Data_scale}). A large-scale dataset will help enrich the expression diversities and speed up the methodology development. (2) The relatively low efficiency of searching over the extremely large expression space, especially considering that a sample-specific model needs to be optimized for each expression.

	\begin{table}[H]
		\caption{The scale of benchmark datasets in SR studies.}
		\label{Data_scale}
		\vskip 0.15in
		\begin{center}
			\begin{tabular}{lccccr}
				\toprule
				Data set & $\#$ samples\\
				\midrule
				Nguyen \cite{uy2011semantically} & 12\\
				BSR \cite{jin2019bayesian} & 6\\
				AI Feynman \cite{udrescu2020ai} & 100\\ 
				SRBench \cite{la2021contemporary} & 252\\
				\textbf{SET-testing} & \textbf{53889}\\
				\bottomrule
			\end{tabular}
		\end{center}
		\vskip -0.1in
	\end{table}

	To address these issues, inspired by the fact that human beings can infer a function based on its curve, we proposed the \textbf{Symbolic Expression Transformer (SET)} from the CV view by representing sampled data as images. SET is the first model to predict symbolic expression from the perspective of CV, to the best of our knowledge. We provide a large-scale dataset generation and separation strategy for training and assessing SR approaches. There is no overlap between training and testing sets in both image domain and equation domains. SET is a sample-agnostic model for SR, so that a trained SET model can be directly applied on testing set. Experimental results demonstrate the promising direction of image-based model for solving the challenging SR problem.

	\section{Methodology}
	\label{Methodology}
	SR aims to find a mapping function from sampled data to symbolic expression. In SET model, the input data are represented as images and the output are symbolic sequences. As a result, we arrive at an image caption problem to translate images to sequences. In this study, we provide the generation strategy of large-scale datasets as well as those details about the proposed SET model.

	\subsection{Dataset generation}
	To generate a large-scale dataset in SR, we need to randomly sample mathematical expressions, draw corresponding images and separate them into training and testing sets without overlap. 
	
	\textbf{Dictionary preparation.}
	In this study, we focus on expressions with no more than two arguments ($x$ and $y$), which can be written as $f(x,y)$. We firstly define a dictionary including 12 operators, 2 variables and 6 constants. The details of dictionary are shown in Table \ref{Dictionary}. Operators may accept one (unary) or two (binary) arguments.
	
	\begin{table}[H]
		\caption{The dictionary for symbolic expression.}
		\label{Dictionary}
		\begin{tabular}{lccccr}
			\toprule
			Variable & \(x,\ y\) \\
			\midrule
			Opr (unary) &
			\(sin,\ cos,\ log,\ sqrt,\ nega,\ inv, \ exp\)\\
			Opr (binary) & \(+ ,\  - ,\ *,\ /,\ power\)\\
			\midrule
			Constant & \(0.5,\ 1,\ 2,\ 3,\ 4,\ 5\) \\
			\bottomrule
		\end{tabular}
	\end{table}

	\textbf{Expression generation.}
	Mathematical expressions are generated in tree structures, where the nodes can be operators, variables and constants \cite{lample2019deep}. The number of operators k is specified and an initial operator is sampled at the beginning. Then an expression can be generated by repeating the following procedures iteratively. 
	\\\hspace*{1em} (1) Create the required number of blank children for the sampled operator. 
	\\\hspace*{1em} (2) Sample the required number of elements to fill the blanks. If the number of operators in current expression reaches k, fill the blanks with variables or constants, then finish this generation. Otherwise utilize at least one operator when filling blanks.
	\\\hspace*{1em} (3) For each sampled operator, go to step (1).
	
\begin{figure}[h]
		\vskip 0.2in
		\begin{center}
			\centerline{\includegraphics[width=\columnwidth]{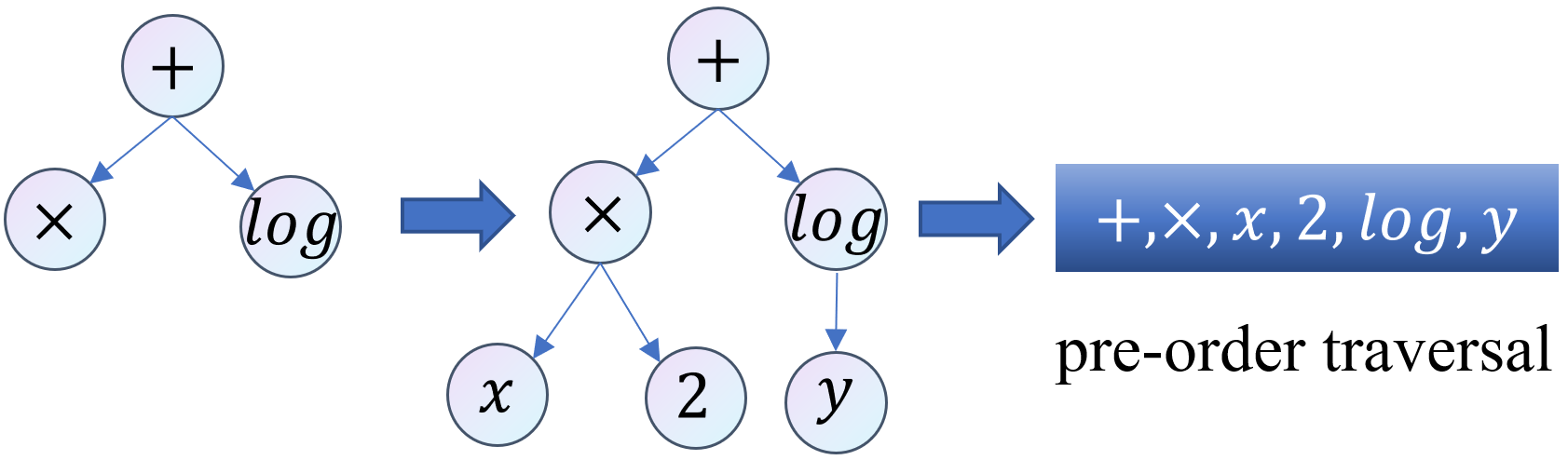}}
			\caption{An example of expression generation: $x*2+log(y)$.}
			\label{generation}
		\end{center}
		\vskip -0.2in
	\end{figure}

	After that, the generated expression is transferred from a tree into a sequence by the pre-order traversal (Figure \ref{generation}).
	
	\textbf{Image construction.}
	To represent comprehensive information within an image, data is collected from different combinations of ranges so that a multi-channel image is constructed for each expression. Each channel is represented by a matrix whose elements are obtained over a specific range according to the following steps: (1) Data sampling from the given range $z_{i,j}=f(x_i,y_j)$; (2) Adding relative noise: $z_{i,j} \leftarrow (1 + 0.01\times \epsilon)z_{i,j}$, where $\epsilon$ is a random value sampled form a Gaussian distribution. (3) Digitizing to $[0, 255]$ by linear mapping; Four-channel images are constructed in this study, where the data are collected from $ \{x\in[L,M],y\in[L,M]\}$, $ \{x\in[L,M],y\in[M,H]\}$, $\{x\in[M,H],y\in[L,M]\}$ and $\{x\in[M,H],y\in[M,H]\}$ respectively, and $L=0.1,M=1,H=7$. 
	
	\begin{figure*}[t]
		\vskip 0.2in
		\begin{center}
			\centerline{\includegraphics[width=1.7\columnwidth]{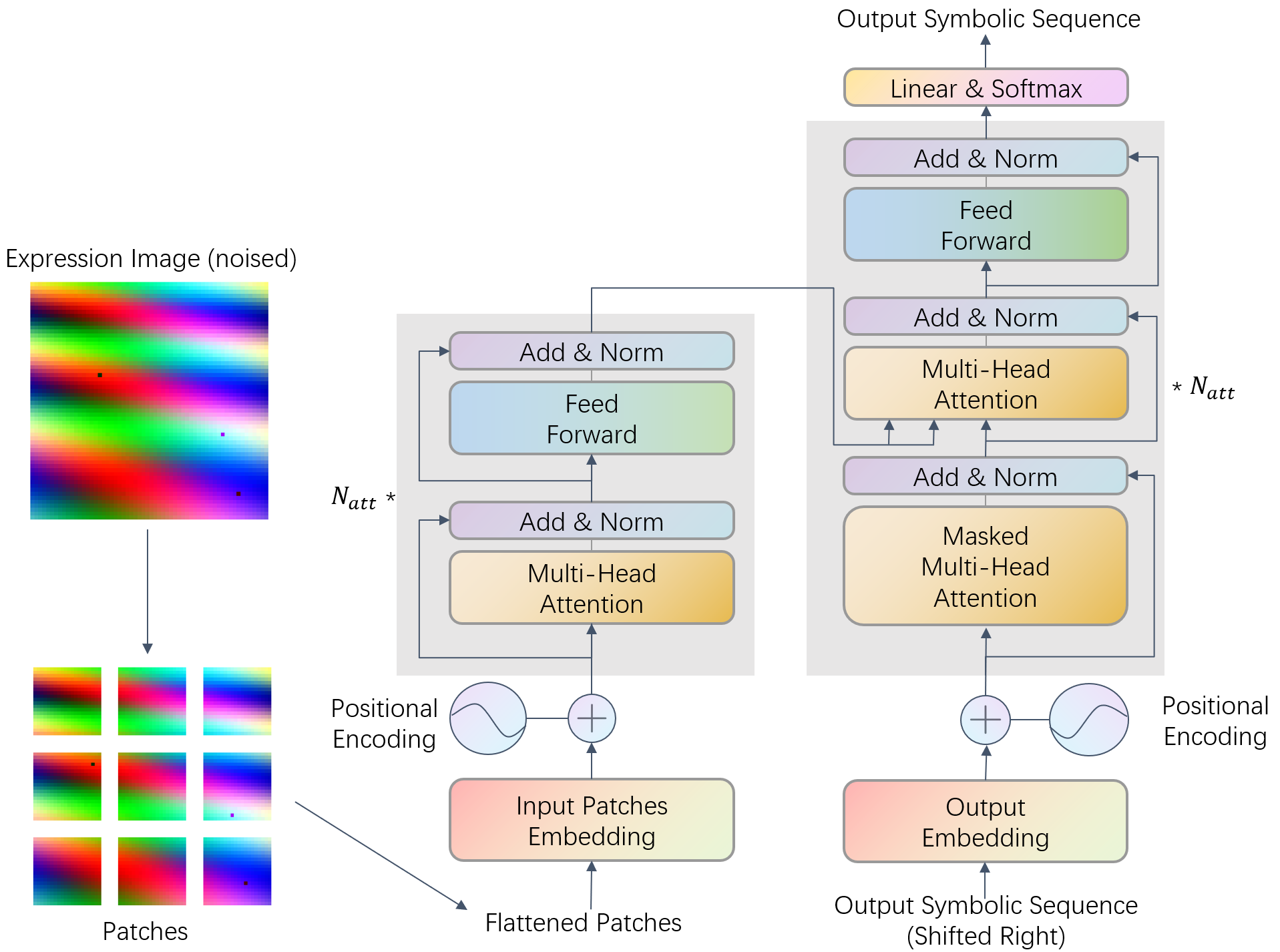}}
			\caption{The structure of SET model. $N_{att}$ is the number of attention blocks. The expression image is cropped into patches and flattened before being fed to the encoder, while the output sequence refers to the pre-order traversal of an expression.}
			\label{model}
		\end{center}
		\vskip -0.2in
	\end{figure*}
	
	\textbf{Training and testing sets separation.}
	There are two principles for expression selection and separation. (1) Expressions associated with the same image should not appear in both training and testing data. (2) Equivalent expressions are desired to be represented in the shorter form. As a result, the dataset is generated in the ascending order respect to the number of operators (from 0 to 6). Each newly sampled expression will be dropped directly if it has the same clean image with a previous sampled one with fewer operators. Additionally, if multiple samples with the same number of operators have the same image, all of them will be saved for improving the diversity of dataset, and then assigned to training or testing data together to avoid overlapping. Considering that the shorter expressions are relatively less and simpler, expressions with fewer than 4 operators will only be assigned to the training set, after being up-sampled to reduce the unbalance within expressions in different lengths. While those expressions with at least 4 operators will be assigned to training set with probability 0.8 and testing set with probability 0.2. Details of the dataset are shown in Table \ref{Augmentation}.
	
	\begin{table}[h]\centering
		\caption{Dataset details with augmentation. To balance the number of samples in different lengths in the training set, expressions with two and three operators are up-sampling to 10 and 2 times respectively.}
		\label{Augmentation}
		\begin{center}
			\begin{tabular}{lccr}
				\toprule	
				$\#$ Operator & Up-sample  & $\#$ Up-sampled train & $\#$ Test \\
				\midrule
				2  & 10 & 34820 & 0 \\
				3  & 2 & 35258 & 0\\
				4  & 1 & 42163 & 10338\\
				5 & 1 & 71065 & 17919\\
				6  & 1 & 102905 & 25632\\
				\textbf{Sum} & -  & \textbf{286211} & \textbf{53889}\\
				\bottomrule
			\end{tabular}	
		\end{center}
	\end{table}
	
	\begin{figure*}[t]
		\vskip 0.2in
		\begin{center}
			\centerline{\includegraphics[width=2\columnwidth]{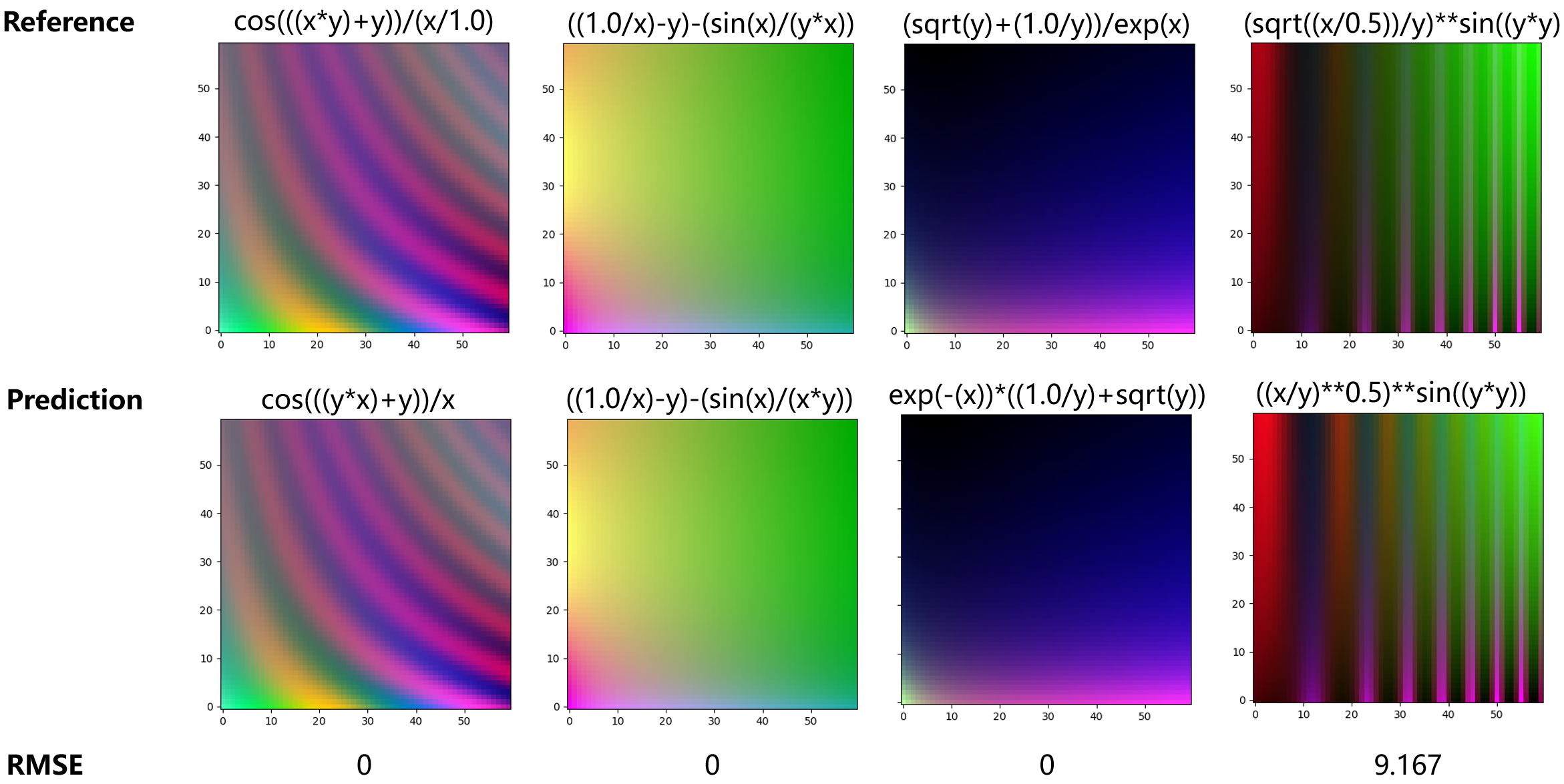}}
			\caption{Visualization of results. The first row shows input expressions with their clean images, while the second row shows the corresponding predictions with their clean images. The RMSE between each pair of images is listed at the bottom.}
			\label{visualization}
		\end{center}
		\vskip -0.2in
	\end{figure*}
	
	\subsection{Model Structure}
	SET models the SR as an image caption task by representing data as images, and applies one-hot embedding to ground truth symbolic sequences. The input to SET is a sequence of cropped image patches while the output is the pre-order traversal of corresponding symbolic expression. Given the marvelous success obtained by transformers \cite{vaswani2017attention} in natural language processing and computer vision, SET further extended the application of transformer models to SR tasks (Figure \ref{model}).

	\textbf{Encoder.}
	In the data loader, Gaussian noise is firstly applied to input images for improving the robustness of trained model. The input image $x\ \in\ R^{H\times W\times C}\ $  is transferred to the patch space $x_p\ \in\ R^{N×\times(P^2 ·C)}$, where $(H,\ W)$ is the shape of the original image, $C$ is the number of channels, $N=(H/P)\ast(W/P)$ is the number of patches and the patch size is $(P,P)$ \cite{dosovitskiy2020image}. Each patch is flattened and reshaped to a one-dimensional vector. As a result, the image is represented by a sequence in length of $N$ when being inputted to the encoder of a transformer. The transformer encoder consists of a linear layer for patch embedding, a positional embedding module, multi-headed self-attention layers and MLP blocks. In each attention head, three weight matrices are learned, which are the query weights $W_Q$, the key weights $W_K$ and the value weights $W_V$. Given the input embedding $E$, a query vector $Q$, a key vector $K$ and a value vector $V$ can be obtained based on those three weights metrics respectively. Specifically, for each token i, $q_i=e_iW_Q, k_i={es}_iW_K, v_i=e_iW_V$. Then the attention weights can be calculated by
	\begin{equation}
	Attention(Q,K,V)\ =\ softmax(\frac{QK^T}{\sqrt{d_k}})V,
	\end{equation}
	where $d_k$ is the dimension of the key vector $K$. Layernorm and residual connections are applied before and after every block, respectively. 
	
	\textbf{Decoder.}
	The decoder basically follows the structure in the standard transformer, which consists of positional encoding, attention blocks, residual connections and layer normalization. The last layer in decoder is a linear transformation with Softmax to output the probability of each character, which refers to the pre-order traversal of an expression. 
	When evaluating a trained model, SET employs the beam search \cite{wu2016google} strategy in the decoder which can provide a set of candidate expressions and avoid getting stuck at local optima. Given an image $I$, the model is supposed to find the sequence $S$ that maximizes a score function $score(S,I)$. A length penalty item $lp(S)$ is added to the score function in testing stage for comparing predicted expressions in different lengths. 
	\begin{equation}
	score\left(S,\ I\right)=\log{\left(P\left(S\middle| I\right)\right)}/lp(S)
	\end{equation}
	\begin{equation}
	lp\left(S\right)=\frac{{(5+|S|)}^\alpha}{{(5+1)}^\alpha}.
	\end{equation}
	
	Among those expressions associated to the same image, SET is supposed to generate the one that being represented in the simplest form. As a result, the $\alpha$ is set to negative values in SET, i.e., $\alpha\ \in(-1,\ 0)$. In out experiment, the beam size is set to 10, which is the number of predicted expressions that will be generated, and the length penalty item $\alpha=-0.9$.

	\section{Results}

	\subsection{Evaluation in image domain}
	In SR tasks, a minor error in the generated sequence can lead to an equation that has definitely different values, or even a symbolic sequence that cannot form an equation. For instance, the pre-order traversals between $-x\ast(sin(x)+cos(y))$ and $exp(x\ast(sin(x)+cos(y)))$ only differ in one item while being the same in seven items, but their value distributions are totally different. On the contrary, sequences which significantly differ from each other may describe functionally the same expression. It is more reasonable to evaluate the predictions in the value space. To maintain the consistency in this study, the performance is assessed by representing values as images again. The image construction strategy is the same with that in dataset generation, while the Gaussian noise is not applied here.
	
	\textbf{Distance in image space.}
	For the $i^{th}$ sample, we can evaluate the image similarity between $I_{hyp}^{i,j}$ and $I_{ref}^i$ by measuring the root-mean-square-error (RMSE) and the structure similarity (SSIM) \cite{wang2004image}, where $I_{hyp}^{i,j}$ is the clean image of the $j^{th}$ prediction in beam size, and $I_{ref}^i$ is the clean image of the reference. We record the minimum RMSE within beam size of each sample, and report the mean value over all samples. The error in image domain is defined as
	\begin{equation}
	RMSE={Mean}_i({Min}_{j\in b e a m}||I_{hyp}^{i,j}-\ I_{ref}^i||_2). \label{RMSE}
	\end{equation}

	Similarly, the SSIM score is defined as:
	\begin{equation}
	S_{SSIM}={Mean}_i({Max}_{j\in beam}(SSIM(I_{hyp}^{i,j},\ I_{ref}^i))). \label{SSIM}
	\end{equation}

	\textbf{Success ratio.}
	SET is supposed to generate expressions who have exact the same value distribution with the corresponding input. If at least one of the predictions within beam search has the same image with the reference, the sample is regarded as being correctly regressed. To be noticed, although a pair of expressions with the same image may not be exactly the same (e.g. $cos(y*x)$ and $cos(-y*x)$), the underlying scientific laws are equivalently between them. Success ratio is defined as:
	\begin{equation}
	R_{succ}=\frac{\# correctly\ regressed\ samples\ }{\# total\ samples}\ast100\%. \label{succ}
	\end{equation}

	To demonstrate the effectiveness of SET model, we make comparison with Nearest Neighbor (NN) algorithm under the same training and testing data. Given the image of a testing sample, NN looks for the expression whose image is most close to the input from the training set. For a fair comparison, the top-10 candidates given by NN are analyzed in the same way as that for the top-10 predictions in beam search of SET. NN needs to calculate the distance between a testing sample to all training samples, which makes it expensive in runtime and memory usage. So the comparison in image space is based on 100 randomly chosen samples (Table \ref{scores}). Other approaches are not included in the comparison for two reasons. (1) The assumptions made in these approaches may not hold in SET dataset. For instance, $sin(x\ +\ cos(x))$ is not allowed in DSR \cite{petersen2021deep} but can exist in SET dataset. (2) The much longer runtime of those searching-based methods limits their feasibility when dealing with a complex expression. For instance, DSR \cite{petersen2021deep} takes up to ${\sim 10}^5$ seconds on an individual sample in our SET dataset. 
	
	\begin{table}[H]
		\caption{Comparison between SET and NN based on top-10 predictions. Considering that there is no repetitive image between training and testing sets, the $R_{succ}$ of NN will always be 0.}
		\label{scores}
		\begin{tabular}{lcr}
			\toprule	
			& NN & SET \\
			\midrule
			RMSE (The lower the better) & 5.66 & 3.55\\
			\(S_{\text{SSIM}}\) (The higher the better) & 0.808 & 0.857\\
			\(R_{\text{succ}}\) (The higher the better) & 0\% & 33\%\\
			Average runtime on each test sample & 11.65s & 1.28s\\
			\bottomrule
		\end{tabular}
	\end{table}

	\subsection{Visualization}
	Results can be visualized by drawing the first three channels of each image. As shown in Figure \ref{visualization}, on the first three samples, SET can find the expression that exactly matches the input image and results in the 0 RMSE. Additionally, referring to the same image, the generated expression can even be simpler than the input one. This is because the negative length penalty item encourages the model to look for expressions with fewer operators. While on the last sample in Figure \ref{visualization}, images of the reference and the prediction are similar but not the same, which is caused by the missing component $/0.5$ in the prediction. Generally speaking, expressions generated by SET always have the similar or even the same value distribution with their corresponding input, which means that SET can find the fundamental relationship across variables and an approximate representation of the desired expression.

	\section{Discussions}
	Inspired by the fact that human beings can infer a mathematical expression based on its function image, we propose SET and a large-scale dataset to address SR from the perspective of computer vision. Experimental results suggest that SET can find predictions whose distribution of values is the same or very close to the desired one, which indicates that the underlying mathematical law between variables are detected from the data.

	Currently, SET model mainly suffers from the weakness in distinguishing expressions with quite similar images. Considering the large scale of SET dataset and the information loss during constructing digital images from the raw data, SET may generate many predictions in the case of the corresponding images differ from each other slightly. In addition, modeling SR as image captioning tasks can only works for functions of no more than two variables, and the training data and testing data need to have the same distribution. Generalization of the model needs to be further explored in the future.

	\bibliographystyle{ieeetr}
	\bibliography{SET}

\begin{thebibliography}{10}

\bibitem{augusto2000symbolic}
D.~A. Augusto and H.~J. Barbosa, ``Symbolic regression via genetic
  programming,'' in {\em Proceedings. Vol. 1. Sixth Brazilian Symposium on
  Neural Networks}, pp.~173--178, IEEE, 2000.

\bibitem{back2018evolutionary}
T.~B{\"a}ck, D.~B. Fogel, and Z.~Michalewicz, {\em Evolutionary computation 1:
  Basic algorithms and operators}.
\newblock CRC press, 2018.

\bibitem{dubvcakova2011eureqa}
R.~Dub{\v{c}}{\'a}kov{\'a}, ``Eureqa: software review,'' 2011.

\bibitem{haeri2017statistical}
M.~A. Haeri, M.~M. Ebadzadeh, and G.~Folino, ``Statistical genetic programming
  for symbolic regression,'' {\em Applied Soft Computing}, vol.~60,
  pp.~447--469, 2017.

\bibitem{koza1992programming}
J.~Koza, ``On the programming of computers by means of natural selection,''
  {\em Genetic programming}, 1992.

\bibitem{schmidt2009distilling}
M.~Schmidt and H.~Lipson, ``Distilling free-form natural laws from experimental
  data,'' {\em science}, vol.~324, no.~5923, pp.~81--85, 2009.

\bibitem{uy2011semantically}
N.~Q. Uy, N.~X. Hoai, M.~O’Neill, R.~I. McKay, and E.~Galv{\'a}n-L{\'o}pez,
  ``Semantically-based crossover in genetic programming: application to
  real-valued symbolic regression,'' {\em Genetic Programming and Evolvable
  Machines}, vol.~12, no.~2, pp.~91--119, 2011.

\bibitem{udrescu2020ai}
S.-M. Udrescu and M.~Tegmark, ``Ai feynman: A physics-inspired method for
  symbolic regression,'' {\em Science Advances}, vol.~6, no.~16, p.~eaay2631,
  2020.

\bibitem{kusner2017grammar}
M.~J. Kusner, B.~Paige, and J.~M. Hern{\'a}ndez-Lobato, ``Grammar variational
  autoencoder,'' in {\em International conference on machine learning},
  pp.~1945--1954, PMLR, 2017.

\bibitem{sahoo2018learning}
S.~Sahoo, C.~Lampert, and G.~Martius, ``Learning equations for extrapolation
  and control,'' in {\em International Conference on Machine Learning},
  pp.~4442--4450, PMLR, 2018.

\bibitem{petersen2021deep}
B.~K. Petersen, M.~L. Larma, T.~N. Mundhenk, C.~P. Santiago, S.~K. Kim, and
  J.~T. Kim, ``Deep symbolic regression: Recovering mathematical expressions
  from data via risk-seeking policy gradients,'' in {\em International
  Conference on Learning Representations}, 2021.

\bibitem{jin2019bayesian}
Y.~Jin, W.~Fu, J.~Kang, J.~Guo, and J.~Guo, ``Bayesian symbolic regression,''
  {\em arXiv preprint arXiv:1910.08892}, 2019.

\bibitem{trujillo2016neat}
L.~Trujillo, L.~Mu{\~n}oz, E.~Galv{\'a}n-L{\'o}pez, and S.~Silva, ``neat
  genetic programming: Controlling bloat naturally,'' {\em Information
  Sciences}, vol.~333, pp.~21--43, 2016.

\bibitem{xing2021automated}
H.~Xing, A.~Salleb-Aouissi, and N.~Verma, ``Automated symbolic law discovery: A
  computer vision approach,'' in {\em Proceedings of the AAAI Conference on
  Artificial Intelligence}, vol.~35, pp.~660--668, 2021.

\bibitem{biggio2021neural}
L.~Biggio, T.~Bendinelli, A.~Neitz, A.~Lucchi, and G.~Parascandolo, ``Neural
  symbolic regression that scales,'' in {\em International Conference on
  Machine Learning}, pp.~936--945, PMLR, 2021.

\bibitem{kamienny2022end}
P.-A. Kamienny, S.~d'Ascoli, G.~Lample, and F.~Charton, ``End-to-end symbolic
  regression with transformers,'' {\em arXiv preprint arXiv:2204.10532}, 2022.

\bibitem{la2021contemporary}
W.~La~Cava, P.~Orzechowski, B.~Burlacu, F.~O. de~Franca, M.~Virgolin, Y.~Jin,
  M.~Kommenda, and J.~H. Moore, ``Contemporary symbolic regression methods and
  their relative performance,'' in {\em Thirty-fifth Conference on Neural
  Information Processing Systems Datasets and Benchmarks Track (Round 1)},
  2021.

\bibitem{lample2019deep}
G.~Lample and F.~Charton, ``Deep learning for symbolic mathematics,'' in {\em
  International Conference on Learning Representations}, 2019.

\bibitem{vaswani2017attention}
A.~Vaswani, N.~Shazeer, N.~Parmar, J.~Uszkoreit, L.~Jones, A.~N. Gomez,
  {\L}.~Kaiser, and I.~Polosukhin, ``Attention is all you need,'' {\em Advances
  in neural information processing systems}, vol.~30, 2017.

\bibitem{dosovitskiy2020image}
A.~Dosovitskiy, L.~Beyer, A.~Kolesnikov, D.~Weissenborn, X.~Zhai,
  T.~Unterthiner, M.~Dehghani, M.~Minderer, G.~Heigold, S.~Gelly, {\em et~al.},
  ``An image is worth 16x16 words: Transformers for image recognition at
  scale,'' in {\em International Conference on Learning Representations}, 2020.

\bibitem{wu2016google}
Y.~Wu, M.~Schuster, Z.~Chen, Q.~V. Le, M.~Norouzi, W.~Macherey, M.~Krikun,
  Y.~Cao, Q.~Gao, K.~Macherey, {\em et~al.}, ``Google's neural machine
  translation system: Bridging the gap between human and machine translation,''
  {\em arXiv preprint arXiv:1609.08144}, 2016.

\bibitem{wang2004image}
Z.~Wang, A.~C. Bovik, H.~R. Sheikh, and E.~P. Simoncelli, ``Image quality
  assessment: from error visibility to structural similarity,'' {\em IEEE
  transactions on image processing}, vol.~13, no.~4, pp.~600--612, 2004.

\end{thebibliography}

	\newpage
	\onecolumn
	\appendix
	
	\begin{algorithm} [h] 
		\caption{Data generation and assignment algorithm}  
		\label{alg}
		\begin{algorithmic}[1] 
			\STATE Specify the minimum and maximum number of operators $n_{min}$ and $n_{max}$. 
			\STATE Initialize expression set $Seq_{all}$, clean image set $I_{all}$, training image set $Img_{train}$,  training expression set $Seq_{train}$, testing image set $Img_{test}$ and testing expression set $Seq_{test}$.
			\FOR {$n$ in $[n_{min}, n_{max}]$}
			\STATE Under current number of operators $n$, initialize training image set $Img^n_{train}$,  training expression set $Seq^n_{train}$, testing image set $Img^n_{test}$,  testing expression set $Seq^n_{test}$, clean training image set $I^n_{train}$ and  clean testing image set $I^n_{test}$.
			\FOR {$i$ in range($t_n$), where $t_n$ is the generation times under current number of operators $n$} 
			\STATE Sample an expression $s_i$ with $n$ operators. 
			\IF {$S_i \in Seq_{all}$,} 
			\STATE Continue.  $\qquad//$ The expression has already been sampled.
			\ELSE {\STATE Add $S_i$ to $Seq_{all}$.}
			\ENDIF  
			
			\STATE Collect data for function $S_i$ over ranges.
			\IF {Value error arises during calculation} 
			\STATE Continue.
			\ENDIF
			
			\STATE Draw the image $I_i$ of expression $S_i$.
			\IF {$I_i \in I_{all}$} 
			\STATE Continue.  $\qquad//$ A shorter expression with the same image has already been sampled.
			\ENDIF
			
			\hspace*{\fill} \\
			$//$ Training and testing set separation.
			\IF {$I_i \in I^n_{train}$} 
			\STATE Flag = 1.
			\ELSIF{$I_i \in I^n_{test}$}
			\STATE Flag = 0.
			\ELSE{
				\STATE  With probability $p$: Flag = 1. Add $I_i$ to $I^n_{train}$.
				\STATE With probability $1-p$: Flag = 0. Add $I_i$ to $I^n_{test}$.}
			\ENDIF
			
			\hspace*{\fill} \\
			\STATE Add Gaussian noise to $I_i$: $Img_i=I_i+\epsilon$ .
			\IF {Flag = 1} 
			\STATE Assign $(Img_i, S_i)$ to $(Img^n_{train}, Seq^n_{train})$.
			\ELSE
			\STATE Assign $(Img_i, S_i)$ to $(Img^n_{test}, Seq^n_{test})$.
			\ENDIF
			\ENDFOR
			\STATE Add $I^n_{train}$ and $ I^n_{test}$ to $I_{all}$.
			\STATE Add $(Img^n_{train}, Seq^n_{train})$ to $(Img_{train}, Seq_{train})$.
			\STATE Add $(Img^n_{test}, Seq^n_{test})$ to $(Img_{test}, Seq_{test})$.
			\ENDFOR
		\end{algorithmic}  
	\end{algorithm}


\end{document}